%% file: iclr2025_conference.tex
\documentclass{article} 
\usepackage{iclr2025_conference,times}

\input{math_commands.tex}

\usepackage{hyperref}
\usepackage{url}
\usepackage[utf8]{inputenc} 
\usepackage[T1]{fontenc}    
\usepackage{hyperref}       
\usepackage{url}            
\usepackage{booktabs}       
\usepackage{amsfonts}       
\usepackage{nicefrac}       
\usepackage{microtype}      
\usepackage{subscript}
\usepackage{colortbl} 
\usepackage{booktabs} 
\definecolor{lightblue}{rgb}{0.68, 0.85, 0.9}
\usepackage{enumitem}
\usepackage{xcolor}
\usepackage{amsmath}
\usepackage{hyperref}
\usepackage{graphicx}
\usepackage{algorithm}
\usepackage{algorithmic}
\usepackage[most]{tcolorbox}
\usepackage{subcaption}

\title{Explain-Query-Test: Self-Evaluating LLMs Via Explanation and Comprehension Discrepancy}

\author{
    Saeid Asgari Taghanaki \\
    Autodesk Research \\
    \And
    Joao Monteiro \\
    Autodesk Research \\
}

\newcommand{\eqt}{\textsc{EQT}}
\newcommand{\mmlupro}{\textsc{MMLU-Pro}}

\iclrfinalcopy 
\begin{document}

\maketitle

\begin{abstract}
Large language models (LLMs) have demonstrated remarkable proficiency in generating detailed and coherent explanations of complex concepts. However, the extent to which these models truly comprehend the concepts they articulate remains unclear.
To assess the level of comprehension of a model relative to the content it generates, we implemented a self-evaluation pipeline where models: (\textbf{i}) given a topic generate an excerpt with information about the topic, (\textbf{ii}) given an excerpt generate question-answer pairs, and finally (\textbf{iii}) given a question generate an answer. We refer to this self-evaluation approach as Explain-Query-Test (\eqt{}).
Interestingly, the accuracy on generated questions resulting from running the \eqt{} pipeline correlates moderatly with the model performance as verified by typical benchmarks such as \mmlupro{}. In other words, \eqt{}'s performance is predictive of \mmlupro{}'s, and \eqt{} can be used to rank models without the need for any external source of evaluation data other than lists of topics of interest.
Moreover, our results reveal a disparity between the models' ability to produce detailed explanations and their performance on questions related to those explanations. This gap highlights fundamental limitations in the internal knowledge representation and reasoning abilities of current LLMs. We release the code at \url{https://github.com/asgsaeid/EQT}.
\end{abstract}


\section{Introduction}

Large language models (LLMs) have achieved remarkable success in natural language processing tasks, including text generation, translation, and question answering \citep{devlin2019bert,brown2020language,openai2023gpt4}. State-of-the-art models such as GPT-4 \citep{openai2023gpt4}, O1-preview, Claude \citep{anthropic2023claude}, Gemini \citep{google2023gemini}, and Llama \citep{touvron2023llama} are capable of producing coherent and detailed explanations about a wide array of concepts.

A critical aspect of intelligence, both human and artificial, lies in the ability to understand and apply knowledge flexibly. While LLMs demonstrate remarkable prowess in generating detailed explanations of concepts, an important question arises: Does this ability reflect true comprehension, or is it simply a sophisticated form of pattern recognition? More specifically, when an LLM explains a concept, can it answer related questions derived from that explanation without direct access to the explanation during testing?

This investigation directly relates to the self-evaluation of LLMs, where models are tasked with assessing their own generated content's alignment with their internal understanding. Self-evaluation is crucial for understanding whether LLMs possess genuine reasoning abilities or merely exploit correlations in training data. By focusing on the relationship between explanation generation and subsequent question answering, we aim to probe the depth of their internal knowledge and the robustness of their reasoning capabilities.

Understanding this disconnect is crucial for several reasons. First, the ability to explain concepts and correctly answer related questions is fundamental for applications in education, healthcare, and decision-making systems \citep{bommasani2021opportunities}. For instance, an LLM used in education should not only provide clear explanations to students but also demonstrate understanding by accurately answering follow-up questions. Second, if models fail at this task, it highlights limitations in their internal knowledge representation and reasoning, signaling risks for high-stakes applications where reliability and understanding are paramount. Finally, this evaluation aligns with broader efforts to ensure that AI systems exhibit true understanding rather than merely leveraging statistical correlations in data \citep{bender2021dangers}.

In this study, we propose a novel self-evaluation framework, Explain-Query-Test (\eqt{}), to assess to what extent state-of-the-art LLMs can independently answer questions derived from their own explanations, without access to those explanations during testing. \eqt{} is performed in three steps: (\textbf{i}) given a topic, a model generates an excerpt with information about the topic, (\textbf{ii}) given an excerpt, the same model then generates question-answer pairs, and finally (\textbf{iii}) a model is given a question and generates an answer.

By decoupling explanation from question answering, \eqt{} tests the models' internal knowledge, reasoning, and consistency, requiring them to rely on deeper comprehension rather than surface-level text patterns. This allows us to rigorously measure not just whether LLMs can generate plausible explanations, but whether they can independently apply their knowledge to novel yet related tasks that revolve around the same underlying knowledge. In Section~\ref{sec:math_theory}, we provide a detailed mathematical justification for the \eqt{} framework, emphasizing its robustness and consistency metrics.

This paper makes the following contributions:
\begin{itemize}
    \item We introduce \eqt{}, a novel framework to assess LLMs' comprehension by evaluating their ability to explain concepts and independently answer questions about their generated explanations.
    \item We propose and analyze metrics, such as answer consistency and stability scores, to evaluate LLMs' ability to reason across paraphrased and conceptually linked questions.
    \item We show that \eqt{} has the potential to be used as a proxy to measure a LLM performance without having a test set.
\end{itemize}


\section{Related Work}

Our work focuses on the self-evaluation of LLMs, specifically on the alignment between their generated explanations and their ability to answer questions derived from those explanations without access to the explanations during questioning. Below, we discuss related work.

\textbf{Explanation Generation in LLMs}.
LLMs have demonstrated the ability to produce detailed and coherent explanations of complex topics, often surpassing human-level articulation in specific domains. Studies such as \citep{wiegreffe2021teachme} and \citep{bansal2022explainability} have focused on the use of LLMs to generate explanations that facilitate understanding and interpretability in downstream tasks. However, these works primarily evaluate the quality of explanations in isolation, without testing whether models can leverage these explanations for reasoning.

\textbf{LLM Question-Answering Capabilities}.
Question-answering (QA) tasks have long been used as benchmarks to evaluate the reasoning and comprehension abilities of LLMs \citep{rajpurkar2016squad, brown2020language}. Recent work highlighted the gap between LLMs' ability to generate correct answers and their performance when faced with nuanced or paraphrased questions \citep{zhao2023paraphrase}, or with questions around knowledge not covered during training \citep{monteiro2024repliqa, monteiro2024xc}. However, little attention has been paid to QA tasks derived directly from the models' own explanations, which is the focus of our study.

\textbf{Self-Evaluation in LLMs}.
Self-evaluation, where LLMs assess or critique their own outputs, has emerged as a promising approach to improve model reliability and reasoning \citep{lightman2023selfeval, madaan2023selfcritique}. For example, \citet{lightman2023selfeval} proposed a framework for models to self-assess their responses, while \citet{madaan2023selfcritique} introduced methods for iterative refinement of model-generated content. \citet{sonoda2024statistical} introduced test data which they used to evaluate variations in textual consistencies across similar content to identify failures in LLMs' self-evaluation. Alternatively, \citet{xia2024language} proposed the use of gaps in likelihood across sequentially obtained generations given a query as means for self-evaluation, but showed that doing so requires an external model to offer a discrepancy reference. Our work complements these studies by examining whether models can demonstrate consistency between their explanations and subsequent reasoning, providing a unique lens on self-evaluation without requiring additional test data or auxiliary models.

\textbf{Evaluating Consistency and Robustness}.
Consistency across paraphrased inputs has been identified as a critical metric for evaluating the robustness of LLMs \cite{wang2023robustness, li2023paraconsistency}. Studies have shown that even state-of-the-art models often fail to maintain stable predictions when faced with semantically equivalent but syntactically varied inputs. We extend this line of inquiry by introducing a consistency metric specific to explanation-derived QA tasks, thereby contributing to the broader effort of evaluating and improving LLM robustness.

Our work bridges the gap between explanation generation, question-answering, and self-evaluation by examining whether LLMs can effectively leverage their own explanations to reason and answer related questions. This approach complements existing research while addressing an underexplored aspect of LLM capabilities.

\section{Explain-Query-Test}

Explain-Query-Test (\eqt{}) evaluates the ability of LLMs to explain concepts and then answer questions based on those explanations. The methodology is designed to work with any set of concepts, making it a flexible framework for assessing LLM performance. We also introduce metrics to measure the performance and consistency of the models across paraphrased questions. \eqt{} is detailed in Algorithm~\ref{alg:concept_evaluation}.

\begin{algorithm}[H]
\caption{Explain-Query-Test}
\label{alg:concept_evaluation}
\begin{algorithmic}[1]
\REQUIRE Set of concepts $C = \{c_1, c_2, \dots, c_n\}$, Language model $LM$, Number of paraphrases per question $k$, Number of questions per concept $q$

\STATE Initialize $Results \leftarrow \{\}$ \COMMENT{Stores results for each concept}
\STATE Initialize $Questions \leftarrow \{\}$ \COMMENT{Stores generated questions}

\FOR{each concept $c \in C$}
    \STATE Prompt $LM$ to provide a detailed explanation $E_c$ for concept $c$
    \STATE Store $E_c$
    \STATE Generate $q$ self-contained multiple-choice questions from $E_c$
    \FOR{each generated question}
        \STATE Create $k$ paraphrased versions of the question
    \ENDFOR
    \STATE Store all questions and paraphrases in $Questions[c]$
\ENDFOR

\FOR{each concept $c \in C$}
    \FOR{each question and its paraphrases in $Questions[c]$}
        \STATE Prompt $LM$ to predict answers for the original and paraphrased questions
        \STATE Record predictions $P$
    \ENDFOR
\ENDFOR

\RETURN $Results$ \COMMENT{Contains Accuracy and Stability for each concept}
\end{algorithmic}
\end{algorithm}

\subsection{Explanation and Question Generation}
For each concept, the LLM is prompted to provide a detailed explanation. From this explanation, multiple self-contained multiple-choice questions are generated. Each question may include multiple correct options. To evaluate the robustness of the model, each question is paraphrased several times, resulting in a set of original and paraphrased questions. These paraphrased questions maintain semantic meaning while varying syntactically.

As will be discussed in Section~\ref{sec:results}, throughout this paper, we use the \mmlupro{} dataset~\citep{wang2024mmlu} to cross-reference results, providing a benchmark for comparison. We note however that \eqt{} can be applied to any set of concepts where explanations and questions can be generated.

\begin{figure}[t!]
\centering
\includegraphics[width=0.98\textwidth]{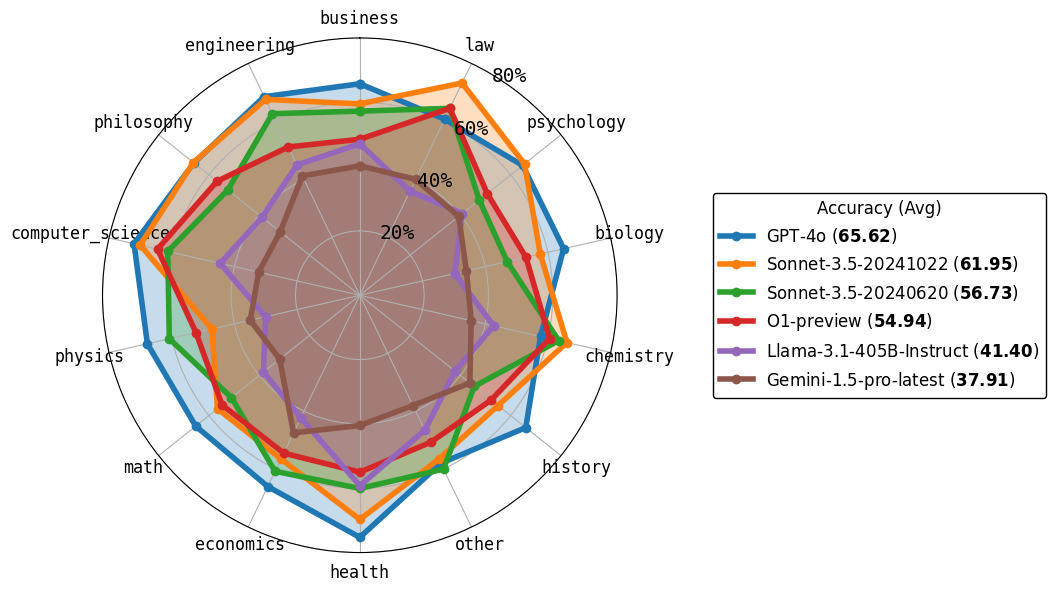}
\caption{Comparison of EQT Accuracy across corresponding MMLU-Pro Categories.}
\label{fig:accuracy-stability-comparison}
\end{figure}

\subsection{A Formal Definition of \eqt{}}
\label{sec:math_theory}

The \eqt{} framework evaluates the reasoning robustness of large language models (LLMs) by testing the consistency and coherence of their internal knowledge representation ($K$). The framework assumes that even if intermediate artifacts, such as explanations ($E$) and questions ($Q$), are partially flawed, the full loop of Explain $\to$ Query $\to$ Test should reflect the model's self-consistent reasoning. This section formalizes the theoretical foundation of the \eqt{} framework.

\subsubsection{Self-Contained Feedback Loop}
In the \eqt{} loop, the model generates:

\begin{enumerate}
    \item Explanations: $E = g_{\theta}(K)$, where $g_{\theta}$ maps the model's internal knowledge representation $K$ to natural language explanations, given a topic.
    \item Questions: $Q = h_{\theta}(E)$, where $h_{\theta}$ transforms explanations into self-contained, logically valid questions.
    \item Answers: $A = f_{\theta}(Q)$, where $f_{\theta}$ represents the reasoning function that maps questions to answers.
\end{enumerate}

Crucially, the \eqt{} framework relies on the level of alignment between the model's predicted answers ($A$) and its internal knowledge representation ($K$), irrespective of inaccuracies in $E$ or $Q$:
\[
    f_{\theta}(h_{\theta}(g_{\theta}(K))) \approx K.
\]

\subsubsection{Full-Loop Accuracy}
The framework evaluates the model's reasoning robustness and self-consistency through full-loop accuracy, defined as:
\[
    \text{Acc}_{\text{loop}} = \frac{\text{Correct Answers Based on Self-Generated } Q}{\text{Total Questions}}.
\]

Accurate models in the $\text{Acc}_{\text{loop}}$ sense can reason consistently within their own generated context, even when intermediate outputs ($E$ and $Q$) are partially flawed (e.g., non-factual or non-grammatical). This metric reflects the internal knowledge consistency of the model.

\subsubsection{Significance of High and Low $\text{Acc}_{\text{loop}}$}

\paragraph{High $\text{Acc}_{\text{loop}}$ Demonstrates Effective Reasoning and Self-Consistency.}
\textit{If $\text{Acc}_{\text{loop}}$ is high, it validates that EQT faithfully captures the model's internal reasoning and consistency.}

Proof:
\begin{enumerate}
    \item Consistency Across Artifacts: The EQT framework ensures that explanations ($E$), questions ($Q$), and answers ($A$) are all derived from the model's internal knowledge representation ($K$). Formally:
    \[
        E = g_{\theta}(K), \quad Q = h_{\theta}(E), \quad A = f_{\theta}(Q).
    \]
    Even if $E$ or $Q$ contain minor flaws, their shared origin in $K$ ensures alignment and consistency.

    \item High \( \text{Acc}_{\text{loop}} \) implies that:
    \[
        f_{\theta}(h_{\theta}(g_{\theta}(K))) \approx K.
    \]
    This reflects that the model's internal knowledge representation is robust and self-consistent, as it can reproduce correct answers despite potential imperfections in intermediate artifacts.

    \item Validation of Reasoning: By isolating the reasoning function $f_{\theta}$ during the Test step, the EQT framework ensures that high $\text{Acc}_{\text{loop}}$ is a measure of the model's internal reasoning rather than reliance on surface patterns or external artifacts.
\end{enumerate}

\paragraph{Low $\text{Acc}_{\text{loop}}$ Indicates Misalignment.}
\textit{If $\text{Acc}_{\text{loop}}$ is low, it highlights deficiencies in the EQT loop, signaling misalignment between generated artifacts ($E$, $Q$) and the internal knowledge representation ($K$).}

Proof:
\begin{enumerate}
    \item Low $\text{Acc}_{\text{loop}}$ implies that:
    \[
        f_{\theta}(h_{\theta}(g_{\theta}(K))) \not\approx K.
    \]
    This discrepancy can arise from:
    \begin{itemize}
        \item Inadequacies in $g_{\theta}$, leading to incomplete or incorrect explanations ($E$).
        \item Flaws in $h_{\theta}$, resulting in poorly constructed or ambiguous questions ($Q$).
        \item Weaknesses in $f_{\theta}$, reflecting limited reasoning or inference capabilities.
    \end{itemize}

    \item Implications for Model Design: Low $\text{Acc}_{\text{loop}}$ reveals areas where the model fails to maintain internal consistency or effectively utilize its knowledge representation. This highlights specific components ($g_{\theta}$, $h_{\theta}$, $f_{\theta}$) that require improvement.
\end{enumerate}

High \( \text{Acc}_{\text{loop}} \) validates the EQT methodology as a robust measure of reasoning, ensuring internal knowledge consistency and alignment across self-generated artifacts. Conversely, low \( \text{Acc}_{\text{loop}} \) acts as a diagnostic tool, pinpointing areas for improvement in the model's reasoning and generation processes. By rigorously testing full-loop accuracy and answer consistency, the \eqt{} framework evaluates reasoning robustness in language models. Through the isolation of the Test step and reliance on self-generated artifacts, \eqt{} ensures that the evaluation reflects the model's internal knowledge representation (\( K \)) and reasoning ability (\( f_{\\theta} \)), independent of the external correctness of explanations or questions.

\section{Experiments}

Experiments are conducted by relying on the set of topics within the widely popular \mmlupro{} dataset, which consists of 14 categories. For each such category, we extract 20 concepts. We then apply \eqt{} for each concept. Questions derived from explanations are compared to the original \mmlupro{} dataset questions to assess the model's ability to adapt to semantically similar yet syntactically varied prompts. 

In further detail, for each category, LLMs generate explanations, and questions are derived based on those explanations. Each question is further paraphrased three times, resulting in a total of 20 questions per concept. This allows us to evaluate the stability of the models across syntactically diverse yet semantically equivalent prompts. The derived questions are compared to the models' known performance on the original \mmlupro{} dataset to analyze consistency and accuracy. The categories in the dataset enable a detailed evaluation of model performance across a wide range of domains.

\begin{figure}[t!]
\centering
\includegraphics[width=\textwidth]{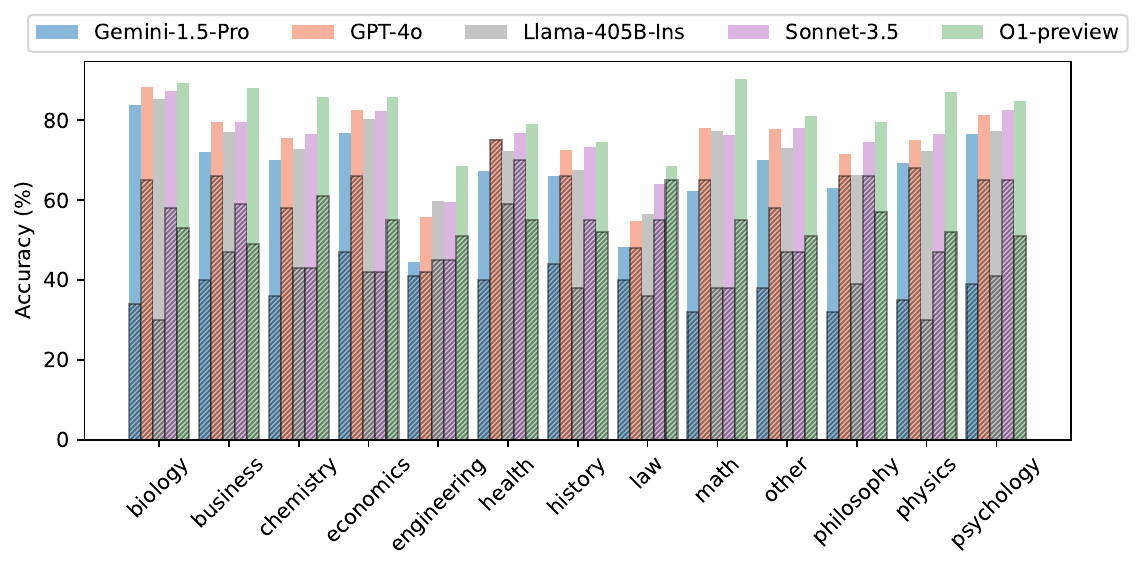}
\caption{Accuracy (\%) comparison across \mmlupro{} categories for various language models. Each model is representedf by two bars: the first (solid) represents the original \mmlupro{} accuracy, and the second (hatched) indicates the adjusted accuracy due to the application of \eqt{} since new questions are added.}
\label{fig:drop_mmlup}
\end{figure}

\subsection{Metrics}
To evaluate the model's ability to answer questions derived from its own explanations, we employ two metrics: question-answer accuracy, and a notion of consistency we define below.

\paragraph{Answer Consistency Score (ACS):}
To measure how stable the model's predictions are across paraphrased questions, we define Answer Consistency Score (ACS). ACS evaluates to what extent the model provides consistent answers when the same question is phrased differently. Note that, in this context, \emph{answers} refer specifically to the option letters (e.g., \( A, B, C \)) selected by the model, rather than the full answer text. Let:

\begin{itemize}
    \item \( A = [a_0, a_1, \dots, a_n] \): The multiset (or list\footnote{A multiset would be a \emph{Counter} in pythonic jargon.}) of predicted answers for a question and its \( n \) paraphrases, where \( a_0 \) is the answer to the original question, and \( a_1, \dots, a_n \) are the answers to its paraphrases.
    \item \( |A| = n + 1 \): The total number of answers.
    \item \( U(A) = \{ a_i \mid a_i \in A \} \): The support of \( A \) given by the set of unique answers in \( A \).
    \item \( |U(A)| \): The number of unique answers in \( A \).
\end{itemize}

ACS is thus defined as:
\begin{equation}
\text{ACS} = 1 - \frac{|U(A)| - 1}{|A|},
\end{equation}
where higher values of ACS indicate greater consistency: \emph{i.e.}, \( \text{ACS} = 1 \) if and only if all answers are identical (maximum consistency), and \( \text{ACS} = 0 \) if and only if all answers are unique (minimum consistency).

This metric highlights areas where the model's behavior is inconsistent, providing insights into the reliability of LLMs when faced with paraphrased versions of questions.


\subsection{Results}
\label{sec:results}

Figure~\ref{fig:drop_mmlup} illustrates the accuracy of various language models on questions derived from the \mmlupro{} categories, as well as the corresponding performance drops compared to the original \mmlupro{} dataset. The solid bars represent the original accuracy on \mmlupro{}, while the hatched bars depict the adjusted accuracy when evaluated using \eqt{}, which reflects the models' concept comprehension.

Interestingly, we observe that models with higher initial accuracy on the original dataset tend to exhibit larger performance drops under \eqt{}. For instance, categories such as \texttt{biology} and \texttt{psychology}, where models initially perform well, show significant degradation in accuracy. On the other hand, categories such as \texttt{law} and \texttt{engineering}, where models already exhibit lower baseline performance, experience smaller relative drops. This trend suggests that the drop in performance is influenced by the disparity between surface-level accuracy and the deeper understanding required to answer questions derived from explanations. Models may struggle to leverage the same high accuracy in the original dataset to maintain consistency in tasks that demand conceptual reasoning.

Overall, these results emphasize the challenges models face when answering questions derived from explanations and underline the need for improved reasoning capabilities. Figure~\ref{fig:spider_plot_accuracy} provides a spider plot visualization of the performance across different categories, offering a detailed view of the model's strengths and weaknesses.

\begin{figure}[t!]
\centering
\begin{subfigure}{.35\textwidth}
    \centering
    \includegraphics[width=\textwidth]{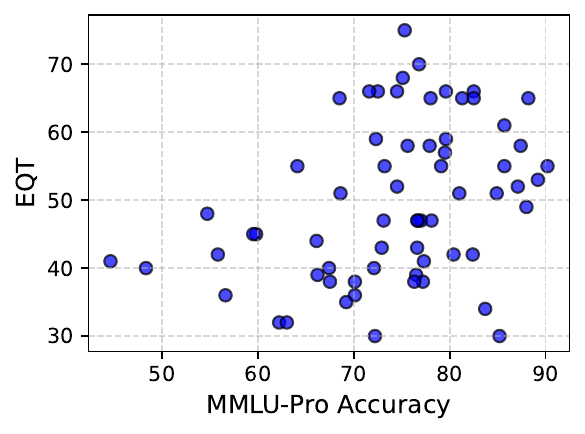}
    \caption{Correlation between MMLU-Pro and EQT. Correlation: 0.361, p-value: 3.151e-03.}
    \label{fig:corr_mmlup}
\end{subfigure}
\hfill
\begin{subfigure}{.6\textwidth}
    \centering
    \includegraphics[width=\textwidth]{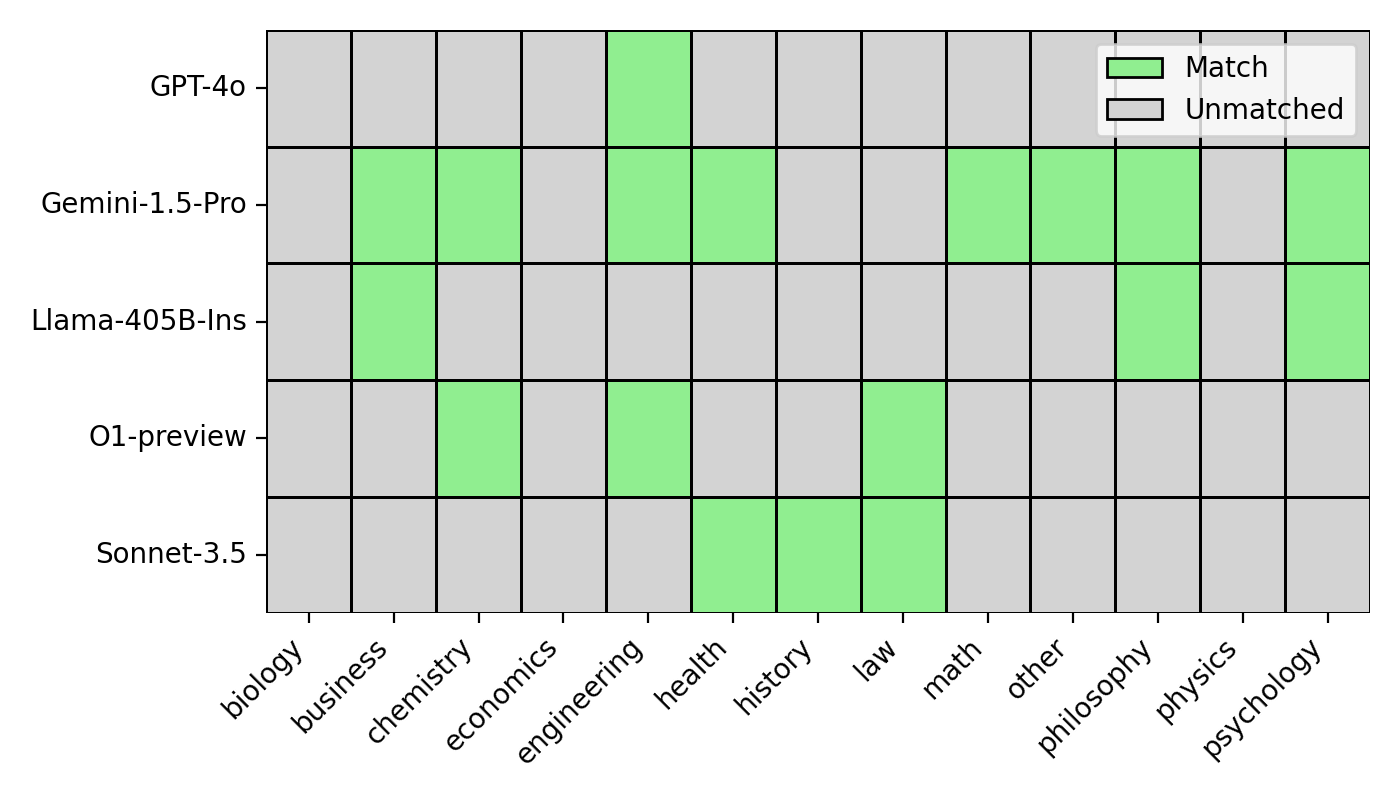}
    \caption{Ranking match heatmap.}
    \label{fig:heatmap_match}
\end{subfigure}
\caption{Analysis of \mmlupro{} and \eqt{} results.}
\label{fig:combined_analysis}
\end{figure}

We further analyzed the relationship between the \mmlupro{} and \eqt{} performances, and results in Figure~\ref{fig:corr_mmlup} show a statistically significant correlation (\(r = 0.361\), \(p = 0.003\)) between the two sets of results. In other words, results are moderately correlated, suggesting that \eqt{} has predictive power of downstream reasoning performance and can serve as a proxy for assessing and ranking the performance of LLM. We highlight that the observed correlation suggests that reasoning-able LLMs can be compared without relying on external test sets, simplifying the evaluation process. By leveraging this proxy approach, we gain insights into LLM performance efficiently and consistently, reducing dependence on extensive datasets while maintaining meaningful performance comparisons.

Additionally, the ranking heatmap displayed in Figure~\ref{fig:heatmap_match} shows the consistency of model rankings across categories for both \mmlupro{} and \eqt{} methods. Each row corresponds to a model, and each column corresponds to a category. Green cells indicate a match in model rankings between the two evaluation methods, while grey cells signify mismatches. Rankings are calculated independently for each method by sorting accuracy scores in descending order, with ties assigned the average rank. The heatmap highlights areas where \eqt{} aligns with the established \mmlupro{} benchmark. Gemini-1.5-Pro has the most matches, while GPT-4o has the fewest, highlighting an interesting variation. Exploring these patterns further on additional benchmark datasets would be an interesting direction for future research.

\subsection{Stability Scores}
To evaluate the consistency of the models' predictions across paraphrased questions, we computed the ACS \textit{stability scores} for each category and model. Table~\ref{tab:stability-scores} summarizes these results. Higher stability scores indicate that the models provided consistent answers despite syntactical variations in the questions.

\begin{table}[t!]
\centering
\caption{ACS stability scores (in $[0-1]$) for each category and model. Higher scores indicate better stability.}
\label{tab:stability-scores}
\resizebox{\textwidth}{!}{%
\begin{tabular}{lccccc}
\toprule
Category & Gemini-1.5-Pro & GPT-4o & Sonnet-3.5 & Llama-405B-Ins & O1-preview \\
\midrule
business & 0.85 & \textbf{0.95} & 0.96 & 0.96 & 0.89 \\
law & 0.85 & \textbf{0.96} & 0.97 & 0.92 & 0.93 \\
psychology & 0.84 & 0.94 & \textbf{0.98} & 0.92 & 0.89 \\
biology & 0.86 & \textbf{0.95} & 0.95 & 0.91 & 0.88 \\
chemistry & 0.87 & 0.96 & \textbf{0.97} & 0.91 & 0.91 \\
history & 0.89 & \textbf{0.97} & 0.96 & 0.92 & 0.92 \\
other & 0.86 & 0.92 & \textbf{0.98} & 0.95 & 0.90 \\
health & 0.86 & 0.96 & \textbf{0.99} & 0.92 & 0.93 \\
economics & 0.87 & 0.95 & \textbf{0.98} & 0.94 & 0.93 \\
math & 0.85 & \textbf{0.97} & 0.96 & 0.91 & 0.89 \\
physics & 0.86 & \textbf{0.95} & 0.94 & 0.88 & 0.90 \\
computer\_science & 0.82 & \textbf{0.97} & 0.97 & 0.93 & 0.93 \\
philosophy & 0.83 & 0.94 & \textbf{0.97} & 0.95 & 0.89 \\
engineering & 0.84 & 0.96 & \textbf{0.98} & 0.91 & 0.92 \\
\midrule
\rowcolor{lightblue}
Average & 0.85 & 0.95 & \textbf{0.97} & 0.92 & 0.91 \\
\bottomrule
\end{tabular}
}
\end{table}

ACS scores indicate that models such as GPT-4o and Sonnet-3.5 consistently outperform others in maintaining answer consistency across paraphrased questions, especially in categories like health and engineering. Figure~\ref{fig:accuracy-stability-comparison} provides a visual comparison of model performance in terms of accuracy and stability scores across the 14 categories in the \mmlupro{}. Overall, GPT-4o demonstrates superior performance compared to other methods, closely followed by the Claude-3.5-Sonnet variants. Interestingly, O1-preview, which is designed to excel in reasoning tasks, falls short of these models.

\section{Conclusion}

In this study, we investigated the extent to which large language models (LLMs) can comprehend and leverage their own explanations to answer related questions. Using the \eqt{} approach we introduced, we evaluated models by prompting them to generate detailed explanations and then testing their ability to answer derived questions independently.

The results revealed a significant gap between the models' ability to generate coherent explanations and their performance on questions derived from those explanations. This discrepancy highlights fundamental limitations in the internal knowledge representation and reasoning capabilities of current LLMs. Furthermore, our evaluation of answer consistency through the Answer Consistency Score (ACS) demonstrated that even state-of-the-art models struggle to maintain consistency across paraphrased questions. Interestingly, we identified \eqt{}'s results have predictive power with respect to \mmlupro{} performance, yielding potential data-less self-evaluation framework for models able to reason.

These findings emphasize the need for further advancements in LLMs to improve their reasoning and understanding. Future work could explore techniques to align explanation generation with robust question-answering capabilities, develop training objectives that prioritize internal consistency, and design benchmarks that better reflect real-world reasoning challenges. By bridging the gap between explanation generation and question-answering, we can make significant strides toward building language models that not only articulate knowledge effectively but also demonstrate a deeper, more reliable understanding of the concepts they explain.

\bibliography{iclr2025_conference}
\bibliographystyle{iclr2025_conference}

\appendix
\section*{Appendix}
\label{appendix:prompts}
\appendix

\section{Implementation Details and Prompt Templates}

This section provides details on how the different steps in our methodology were implemented, along with the exact prompt templates used for explanation generation, question generation, and question paraphrasing. These prompts were carefully designed to ensure clarity, completeness, and consistency in the tasks performed by the language models (LLMs).

\textbf{Explanation Generation}. To obtain detailed explanations of concepts, we used the following prompt template:

\begin{tcolorbox}[
    sharp corners,
    colback=gray!5,
    colframe=gray!75!black,
    fonttitle=\bfseries,
    title=Prompt for Explanation Generation,
    boxrule=0.8mm,
    width=\textwidth
]

Please provide a comprehensive and detailed explanation of the concept '\{concept\}', including its background, key principles, applications, and examples. Ensure the explanation is thorough and covers all essential aspects.

\end{tcolorbox}

This prompt ensures that the LLM provides a complete and self-contained explanation of the given concept.

\textbf{Question Generation.} For generating questions from the explanations, the following prompt template was used:

\begin{tcolorbox}[
    sharp corners,
    colback=gray!5,
    colframe=gray!75!black,
    fonttitle=\bfseries,
    title=Prompt for Question Generation,
    boxrule=0.8mm,
    width=\textwidth
]

Create a multiple-choice question about the following concept: [CONCEPT]. Use this question type: "\{question type\}"

\textbf{Requirements:}
1. The question should have \{num\_options\} options (A to \{last\_choice\}).
2. There should one or several correct answers.
3. Base the question and all options ONLY on the information provided below.
4. Make the question entirely self-contained. Do NOT refer to any explanation, provided information, or external context.
5. Avoid phrases like "according to the text", "as described", or any similar references.
6. Ensure the question and options are clear and complete on their own.

\textbf{Format Example:}
Question: [Your question here] \\
Options: \\
A) Option A text \\
B) Option B text \\
... \\
Correct Answers: [List of correct option letters, e.g., A, C, F]

Information about [CONCEPT]: \{explanation\}

\end{tcolorbox}

This prompt ensures that the generated questions are accurate, comprehensive, and fully independent of the source material while remaining aligned with the concept's explanation.

\textbf{Question Paraphrasing.} To generate paraphrased versions of the questions, the following prompt was used:

\begin{tcolorbox}[
    sharp corners,
    colback=gray!5,
    colframe=gray!75!black,
    fonttitle=\bfseries,
    title=Prompt for Question Paraphrasing,
    boxrule=0.8mm,
    width=\textwidth
]

Paraphrase the following question without changing its meaning. Ensure the paraphrased question is self-contained and does not reference any previous explanation or use phrases like "as mentioned earlier". ONLY generate the paraphrased question itself, and do not include any extra text such as "Here's a paraphrased version of the question:" or similar.

'\{question\_text\}'

\end{tcolorbox}

This prompt ensures that the paraphrased questions retained their original meaning while introducing syntactic variations.

\textbf{Question Answering.} Finally, for answering the questions, we used the following prompt:

\begin{tcolorbox}[
    sharp corners,
    colback=gray!5,
    colframe=gray!75!black,
    fonttitle=\bfseries,
    title=Prompt for Question Answering,
    boxrule=0.8mm,
    width=\textwidth
]

\{question\}

Options: \\
A) Option A text \\
B) Option B text \\
...

Please select all correct options (e.g., A, C, D) and provide your answer in the format: "Answer: [Your selections]".

\end{tcolorbox}

This prompt instructs the LLM to provide answers in a clear format while respecting the requirements for multiple correct options.

\textbf{Summary of Design Choices}. Each prompt was crafted with specific guidelines to ensure:
\begin{itemize}
    \item \textbf{Completeness:} All outputs are self-contained and do not rely on external context.
    \item \textbf{Clarity:} Prompts clearly define the task and expected format.
    \item \textbf{Robustness:} Iterative refinements were applied to address ambiguities or errors in generated content.
\end{itemize}

These carefully designed prompts form the backbone of our methodology and enable the evaluation of LLMs' capabilities in explanation generation, question-answering, and consistency.

\section{Sample E, Q, and A}

Here is a sample explanation, questions, and paraphrases along with correct answers and predictions provided by GPT-4o for concept \texttt{graph theory} for category \texttt{math}.

\begin{tcolorbox}[
    title= Math: Graph Theory,
    colframe=black,
    colback=white,
    sharp corners,
    breakable 
]
\textbf{Explanation:}  
Graph theory is a branch of mathematics that studies the properties and applications of graphs, which are mathematical structures used to model pairwise relations between objects. It has become a fundamental area of study in discrete mathematics and has applications across various fields, including computer science, biology, social sciences, and more.

\textbf{Background:}  
The origins of graph theory can be traced back to the 18th century with the work of Leonhard Euler. Euler's solution to the Königsberg Bridge Problem in 1736 is often considered the first theorem of graph theory. The problem involved finding a walk through the city of Königsberg that would cross each of its seven bridges exactly once. Euler abstracted the problem into a graph, where landmasses were vertices and bridges were edges, leading to the formulation of Eulerian paths and circuits.

\textbf{Key Principles:}
\begin{itemize}
    \item \textbf{Graphs and Their Components:}
    \begin{itemize}
        \item \textbf{Vertices (or Nodes):} Fundamental units or points in a graph.
        \item \textbf{Edges (or Links):} Connections between pairs of vertices.
        \item \textbf{Directed vs. Undirected Graphs:} Edges in directed graphs indicate one-way relationships, while undirected graphs represent two-way relationships.
    \end{itemize}
    \item \textbf{Types of Graphs:}
    \begin{itemize}
        \item \textbf{Simple Graphs:} No loops or multiple edges between the same pair of vertices.
        \item \textbf{Multigraphs:} Allow multiple edges between the same pair of vertices.
        \item \textbf{Weighted Graphs:} Edges have weights representing costs, distances, or other metrics.
        \item \textbf{Bipartite Graphs:} Vertices can be divided into two disjoint sets such that every edge connects a vertex in one set to a vertex in the other.
    \end{itemize}
    \item \textbf{Graph Properties:}
    \begin{itemize}
        \item \textbf{Degree:} Number of edges incident to a vertex.
        \item \textbf{Path:} Sequence of edges connecting a sequence of vertices.
        \item \textbf{Cycle:} Path that starts and ends at the same vertex without repeating edges or vertices.
        \item \textbf{Connectedness:} Graph is connected if there is a path between any pair of vertices.
    \end{itemize}
\end{itemize}

\vspace{0.5em}

\textbf{Original Question:}  
Explain the key characteristics of graph theory as a mathematical concept.

\textbf{Options:}
\begin{enumerate}[label=\Alph*)]
    \item Graph theory involves the study of vertices and edges to model relationships between objects.
    \item It exclusively focuses on weighted graphs where edges represent costs or distances.
    \item Graph theory originated with the solution to the Königsberg Bridge Problem by Euler.
    \item It includes the study of complete graphs, where each pair of vertices is connected by an edge.
    \item Directed graphs in graph theory have edges that indicate a two-way relationship.
    \item Graph theory is primarily used in biology and has limited applications in computer science.
    \item Concepts like cycles and paths are fundamental to understanding graph connectedness.
    \item Graph theory does not consider the use of algorithms for exploring graph structures.
    \item Multigraphs in graph theory can have multiple edges between the same pair of vertices.
    \item It is a modern mathematical field developed in the late 20th century.
\end{enumerate}

\vspace{0.5em}

\textbf{Paraphrased Questions:}
\begin{enumerate}
    \item What are the key features of graph theory, and how is it defined as a branch of mathematics?
    \item Can you describe the essential characteristics of graph theory and its fundamental principles?
    \item How would you explain graph theory as a mathematical framework for understanding relationships and connections?
\end{enumerate}

\textbf{Correct Answers:}  
A, C, D, G, I.  

\textbf{Predicted Answers by GPT-4o:}  
A, C, D, G, I.
\end{tcolorbox}

\end{document}

%% file: math_commands.tex
\usepackage{amsmath,amsfonts,bm}

\def\eqref#1{equation~\ref{#1}}

\def\1{\bm{1}}

\DeclareMathAlphabet{\mathsfit}{\encodingdefault}{\sfdefault}{m}{sl}
\SetMathAlphabet{\mathsfit}{bold}{\encodingdefault}{\sfdefault}{bx}{n}

